\documentclass{article}

\usepackage[preprint]{nips_2018}

\usepackage[utf8]{inputenc} 
\usepackage[T1]{fontenc}    
\usepackage{hyperref}       
\usepackage{url}            
\usepackage{booktabs}       
\usepackage{amsfonts}       
\usepackage{nicefrac}       
\usepackage{microtype}      
\usepackage{graphicx}
\title{AgentBuddy: A Contextual Bandit based Decision Support System for  Customer Support Agents}

\author{
 Hrishikesh Ganu
  \texttt{hrishikeshvganu@gmail.com} \\
}

\begin{document}

\maketitle

\begin{abstract}
In this short paper, we present early insights from a Decision Support System for Customer Support Agents (CSAs) serving customers of a leading accounting software. The system is under development and is designed to provide suggestions to  CSAs  to make them more productive. A unique aspect of the solution is the use of bandit algorithms to create a tractable human-in-the-loop system that can learn from CSAs in an online fashion. In addition to discussing the ML aspects, we also bring out important insights we gleaned from early feedback from CSAs. These insights motivate our future work and also might be of wider interest to ML practitioners.
\end{abstract}

\section{Introduction}
While ML systems have come a long way in terms of their performance on specific tasks like object detection etc., their performance on NLP tasks lags behind human level performance by a large margin. 


Our current work is in the spirit of taking a pragmatic view, slightly offbeat from the hype around AI. It is inline with the opinion expressed by Michael Jordan  (\citet{jordan_2018}) and others that what we need right now is Intelligence Augmentation (IA) rather than AI. In the following sections we describe a human-in-the-loop ML system based on bandit algorithms which interacts with CSAs and gets ``rewards'' from them. In Section~\ref{ssec:motivation} we describe how we arrived at the problem and some background details. Therein we also list out our main contributions (Section~\ref{ssec:main_contrib}). This is followed by Section~\ref{sec:approach} where we discuss the approach and the architecture. In Section~\ref{sec:results} we describe some early results/insights from trials of the system with CSAs. Then we discuss learning from this initial work in Section \ref{sec:discussion}. Since this paper is about a in-progress system, we also provide our thoughts about  proposed enhancements in Section~\ref{ssec:proposed_enhamcements}. Finally, we end with conclusions.

\subsection{Background}\label{ssec:motivation}
We are a 34 year old company which sells tax and accounting software.  It has products for individuals, self- employed and small business owners. 

Because of a large customer base and versatile product features, our Customer Support Agents (CSAs) receive a huge volume of customer queries via telephone and webchat. The overall goal for CSAs is correct and quick resolution of queries to improve end user experience. 

\subsection{Problem Statement}
We are a Machine Learning team which works closely with the Customer Support team. For the curent paper we focus on webchat as the channel for customer support \footnote{In webchat, customers can type their queries in a chat window with the agent and the agents can answer the questions by typing in their responses}. The overall setup was the following:
\begin{itemize}
\item The chat starts with a customer typing in their query/problem
  \item CSAs try to solve the problem using a combination of A) their own domain knowledge and B) searching through a knowledge base which contains documents about Quickbooks usage and troubleshooting. 
   \item The chat can go into several rounds of back and forth utterances between the customer and the CSA
  
\end{itemize}
With the recent focus on Data Science at our company, internally several teams have developed question-answering systems/models. However most of these systems have been trained in a supervised learning framework for mostly information retrieval tasks either through manual annotations or through other means.  Given such a plethora of systems,  most of which are trained on tasks and optimized for metrics  not directly related to helping the CSA answer questions and given that CSAs don't have time to provide ``supervised''\footnote{By this we mean CSAs don't have time to look at the output from all systems and say which one is the best} feedback we faced some key questions which we highlight in the next section.
\subsection{Key Questions}
\label{ssec:key_questions}
\begin{itemize}
    \item How can we create a ``selector'' mechanism which can leverage these existing models as candidate models for answering the question and helps choose the right\footnote{By right system we mean a system which can provide the best response to the question}.  model, given a question?
    \item How to do this given that we have time to show only one answer and hence observe feedback only for the model which generated that answer? 
\end{itemize}
\subsection{Main contributions}
\label{ssec:main_contrib}
While we do not claim any novel contributions on the algorithmic aspects of bandit models, our key contribution in this early version is primarily around the practical aspects of leveraging bandit algorithms in a real-world system. While Bandit algorithms have a history in academic literature, we are not aware of many  commercial systems that  leverages Bandit algorithms  to help CSAs to answer long tailed questions, specifically in the accounting domain. Another unique aspect of our system is that it interacts with and receives feedback from multiple categories of human users: customers and CSAs.

\section{Approach}
\label{sec:approach}


Based on the considerations, in Section~\ref{ssec:key_questions}, a  reinforcement learning  approach seemed to be a natural choice. 
However given the practical difficulty of learning RL models 
 we chose the Contextual Bandits (CB) framework which is computationally more tractable \footnote{This is more of a folklore statement and we don't attempt to give a rigorous justification around why CB is more tractable than RL}. For this work we experimented with the class of algorithms discussed by ~\citet{lang2018} with implementations available in the software  \citet{vowpalwabbit}.
 \subsection{Mapping to Contextual Bandits Problem}
 Our problem maps to the CB setup thus:
 \begin{enumerate}
     \item The customer's question along with the chat history in the current 
session acts as the context. At each step, the CB model observes the context
\item Each of the internally built models for question-answering related tasks becomes an arm of the bandit. There are around 10 such models.At each step, the CB model chooses one of these 10 arms
\item The chosen model is used to provide the answer to the CSA who provides feedback on a numeric scale of 1-5 (5=best)to the CB model. This feedback is used for updating the weights in an online fashion.

 \end{enumerate}

\label{ssec:architecture}

\subsection{Architecture} 
\begin{figure}
  \centering 
\includegraphics[width=0.5\linewidth]{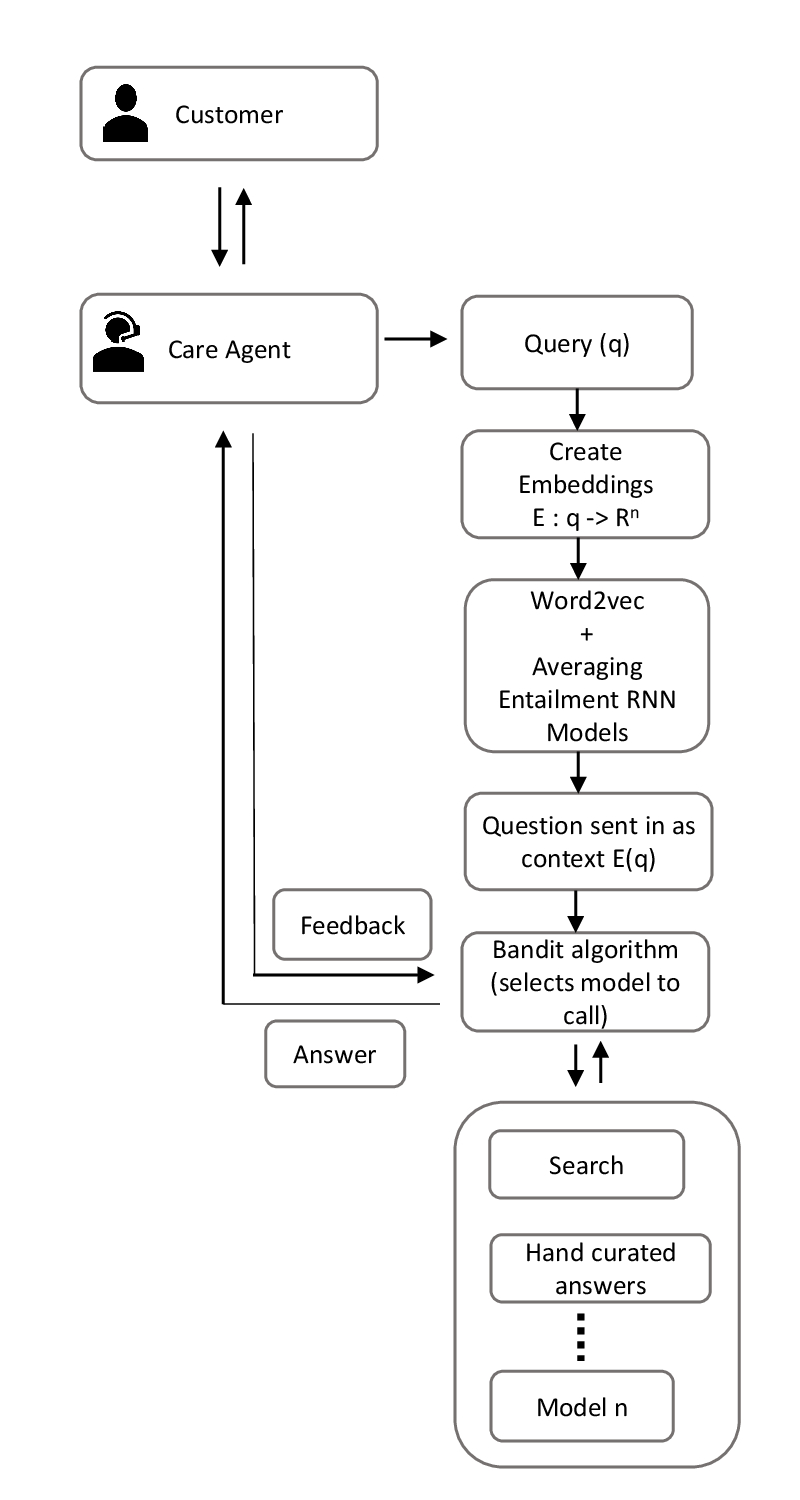}
  \caption{a) The Bandit algorithm chooses one of the models (``arm'' n Bandit terminology) among ``Search'', ``Hand curated answers'' etc. b)The chosen model provides the answer which is observed by the CSA (care Agent) c) The CSA provides feedback on a scale of 1-5 to the Bandit algorithm which updates it's weights based on this feedback. Note that the query coming from the end-customer is mapped to an embedding in $R^{n}$ before it is sent to the Bandit algorithm. Also note that the ML system does not directly provide answers to the end-customer but does so only through the Care Agent (CSA)}
\label{fig_archi}

\end{figure}

Figure ~\ref{fig_archi} shows the architecture of the human-in-the-loop system used to help CSAs. The core of this architecture is the Care agent (CSA): Bandit algorithm setup which makes this a human-in-the-loop system learning in an online fashion.
On the engineering side, the entire setup (except the individual models like ``Search'' which are hosted by several different teams) is on an AWS EC2 machine and all modules have their own Docker containers. This makes the setup modular and easy to maintain. The Care Agents (Customer Support Agents) are provided with an intuitive user interface based on Angular JS on their laptops. They are provide secure and authenticated access to the backend ML system on AWS.

\begin{figure}[htbp]
\centerline{\includegraphics[width=1.2\linewidth]{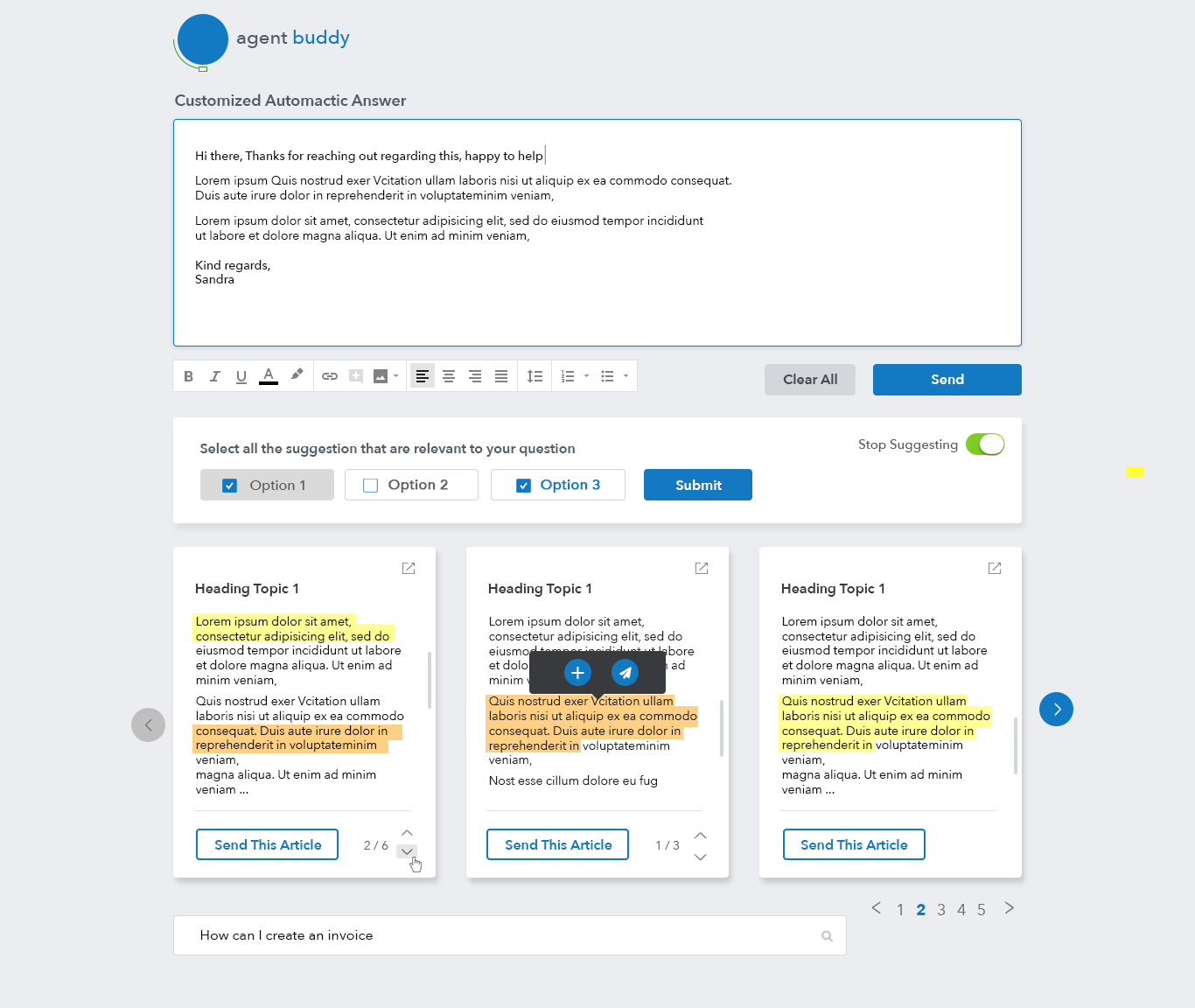}}
\caption{This is the UI for CSAs. In this example, the Bandit algorithm chose ``Search'' as the best arm. The top 3 documents returned by search are shown (only partial documents are shown). CSAs can choose to a)directly send the highlighted snippet to the customer, b)use the snippet for composing their own answer, c)highlight a different section of text of their choice or d)ignore the answer. We record the CSA's choices and this is used as implicit feedback by the bandit model. Note that in addition to just displaying the documents, in each document we highlight text relevant to the question using models trained for Machine Comprehension. This saves precious time for human agents to get to the point and provide superior user experience. This example also shows that our search system can handle spelling mistakes ("supplier" as "suplier").}
\label{fig_ui}
\end{figure}

\section{Results} 
\label{sec:results}
Since this is a system under development we do not have robust quantitative results yet. However, we  provided this in-progress system to some CS agents and asked for feedback. Overall CSAs said they liked the idea of automatically getting suggestions. They also felt that without this system it'd have been tricky for them to look at 10 different responses from different systems to formulate their own answer-all of this while chatting with customers over webchat.

\subsection{Handling ambiguous questions}
\label{ssec:ambiguous_questions}
 One of the key inputs that the CSAs provided
was that the solution seemed to be answering questions prematurely.  To be specific, the models seemed to be answering  even when the questions seemed to be ambiguous.
An example of such a question is \textit{"How can I receive payment?"}. Without more information about which account should be credited (if a user has multiple accounts), mode of payment (wire transfer, cheque etc.) and other attributes it is not possible to give a useful answer. Faced with such an ambiguous question, CSAs felt they would ask a clarifying question like:
\textit{"Do you want to receive payment through wire transfer into your  account \# 12345?"}. CSAs felt that the models should also, like humans ask clarifying questions that can be
sent on to customers for additional information. We build upon this point  in some detail in Section ~\ref{ssec:proposed_enhamcements}. 
\section{Discussion }
\label{sec:discussion}
We have presented a work-in-progress project for a human-in-the-loop system for helping CS agents. The solution is  a central ML platform based on Contextual Bandit algorithms and is tailored to the needs
of the CSA team while enabling reuse of existing models developed in other teams. While the qualitative feedback from CSAs indicates that they see benefits after using the early version, they also brought up the issue of handling ambiguous questions. We discuss in Section \ref{ssec:proposed_enhamcements} some of our early ideas on how to approach this.
\subsection{Limitations} 
Since this is a practical application requiring us to operate under  engineering and organizational constraints the current study has some shortcomings . The following is a list of key  limitations (theoretical and practice wise):
\begin{itemize}
  \item Since the same set of CS agents repeatedly interacts with our system, the future rewards depend on the actions chosen by the algorithm in previous steps. Hence this is not truly a
    contextual bandits setup. 
  \item The features that comprise the context are not very rich currently. E.g: we do not include features from historical chats of customers etc. We plan to enrich this feature set in future.
\end{itemize}
\subsection{Proposed Enhancements}
\label{ssec:proposed_enhamcements}
We discussed in ~\ref{ssec:ambiguous_questions} about the problem with ambiguous questions. If we restrict ourselves to models which provide answers from a fixed set, one of the ways to approach this problem is to assume that the answer to the customer's question exists in the
set of answers available with us.  If we also assume that each 
document in our corpus is   a bag of words, then we can eliminate answers on the basis of words that they contain/do not contain. These additional filters can be provided by the customer under the assumption that the customer has clarity about what they want. \footnote{E.g.: Using the example in Section ~\ref{ssec:ambiguous_questions} we can ask the customer if they want payment through ``internet banking''. If they say yes, then answers which talk about other modes of receiving payment can be discarded.} . In this case we would want to ask a clarifying question in the form of a ``filter'' which would lead to the largest expected reduction in the number of candidate answers remaining. Depending on the peculiarities of the problem, greedy solutions with  good optimality bounds might be available, following the work in \citet{DBLP:journals/jair/GolovinK11}. We wish to pursue this line in future.

\section{Conclusions}
We have described a real life human-in-the-loop system which learns from CSAs through Bandit feedback. We believe that using a Contextual Bandit setup early on  will help us not only to satisfy the needs of CSAs who can provide only  partial feedback but will also help us be prepared for the future when the system might start interacting directly with customers. The ability to learn from partial feedback and in an online fashion make this system attractive from a real world perspective.  Early feedback from CSAs  indicates that even the work-in-progress version is useful to them. Further their feedback has motivated an interesting problem of handling ambiguous questions which can prove to be of wider interest to ML practitioners working on human-in-the-loop systems.

\bibliography{biblio}







\end{document}